\documentclass{sig-alternate}

\usepackage{algorithm}
\usepackage{algpseudocode}
\usepackage{url}
\usepackage{epstopdf}

\begin{document}
%

\newcommand{\TODO}[1]{\textbf{XXX: #1}}
\newcommand{\captionskip}{\vskip -0.15in}

\title{
Scalable methods for nonnegative matrix factorizations of
near-separable tall-and-skinny matrices
}
%
%
%
\numberofauthors{3} 
%
\author{
%
\alignauthor
\mbox{\hspace{-3ex}Austin R.~Benson, Jason D.~Lee}\\
       \affaddr{Stanford University}\\
       \affaddr{Institute for Computational and \\ Mathematical Engineering}\\
       \email{\mbox{\hspace{-2ex}\{arbenson,jdl17\}@stanford.edu}}
\alignauthor
Bartek Rajwa\\
       \affaddr{Purdue University}\\
       \affaddr{Bindley Biosciences Center}\\
       \email{brajwa@purdue.edu}
\alignauthor
David F.~Gleich\\
       \affaddr{Purdue University}\\
       \affaddr{Computer Science}\\
       \email{dgleich@purdue.edu}
}

\maketitle
\begin{abstract}
Numerous algorithms are used for nonnegative matrix
factorization under the assumption that the matrix is nearly
separable.
In this paper, we show how to make these algorithms efficient for
data matrices that have many more rows than columns, so-called
``tall-and-skinny matrices".
One key component to these improved methods is
an orthogonal matrix transformation that preserves the separability of
the NMF problem. Our final methods need a single pass over the
data matrix and are suitable for streaming, multi-core, and MapReduce
architectures.  We demonstrate the efficacy of these algorithms on
terabyte-sized synthetic matrices and real-world matrices from
scientific computing and bioinformatics.
\end{abstract}




\section{Nonnegative matrix factorizations at scale}

A nonnegative matrix factorization (NMF) for an $m \times n$ matrix $X$ with real-valued, 
nonnegative entries is
\begin{equation}
\label{eq:nmf}
X = WH
\end{equation}
where $W$ is $m \times r$, $H$ is $r \times n$, $r < \min(m, n)$, and both factors have nonnegative entries. While there 
are already standard dimension reduction techniques for general matrices such as the singular value decomposition and the interpolative decomposition \cite{cheng2005compression},
the advantage of NMF is in \emph{interpretability} of the data.
A common example is facial image decomposition \cite{guillamet2002non}.
If the columns of $X$ are pixels of a facial image, the columns of $W$ may be facial features such as eyes or ears, and
the coefficients in $H$ represent the intensity of these features.
For this reason, among a host of other reasons, NMF is used in a broad range of applications including graph clustering \cite{kuang2012symmetric}, protein sequence motif discovery \cite{kim2011sparse}, and hyperspectral unmixing \cite{jia2009constrained}.

An important property of matrices in these applications and other massive scientific data sets is that they have many more rows than columns ($m \gg n$).
For example, this matrix structure is common in big data applications with hundreds of millions of samples and a small set of features---see, e.g., Section~\ref{sec:flow} for a bioinformatics application where the data matrix has 1.6 billion rows and 25 columns.
We call matrices with many more rows than columns \emph{tall-and-skinny}.

In this paper, we are concerned with efficient NMF algorithms for tall-and-skinny matrices as
prior work has not taken advantage of this structure for large-scale factorizations.
First, in Section~\ref{sec:dimension_reduction}, we present a new dimension reduction technique using orthogonal transformations.
These transformations are particularly effective for tall-and-skinny matrices
and lead to algorithms that need only one pass over the data matrix.
We compare this method with a Gaussian transformation technique from the hyperspectral unmixing community \cite{bioucas2011overview, boardman1993automating}.
Since tall-and-skinny matrices are amenable to streaming computations \cite{benson2013direct, constantine2011tall},
we present a MapReduce implementation in Section~\ref{sec:implementation}.
However, the algorithms are easily translated to GPUs and distributed memory MPI, since we just need the TSQR kernel (see the discussion in Section~\ref{sec:tsqr}).
We begin by showing that our method correctly recovers the answer hidden in synthetically generated problems in Section~\ref{sec:synthetic}; and then we test our algorithms on data sets from two scientific applications,
heat transfer simulations and flow cytometry, in Section~\ref{sec:applications}.

The software for our algorithms and experiments is available online at \url{https://github.com/arbenson/mrnmf}.

In the remainder of the introduction, we review the state of the art for computing non-negative matrix factorizations.

\subsection{Separable NMF}

We first turn to the issue of how to practically compute the factorization in Equation~(\ref{eq:nmf}).
Unfortunately, for a fixed non-negative rank $r$, finding the factors $W$ and $H$ for which the residual $\| X - WH \|$ is minimized is NP-complete \cite{vavasis2010complexity}.
To make the problem tractable, we make assumptions about the data.
In particular, we require a separability condition on the matrix.
A nonnegative matrix $X$ is \emph{$r$-separable} if
\[
X = X(:, \mathcal{K})H,
\]
where $\mathcal{K}$ is an index set with $|\mathcal{K}| = r$ and $X(:, \mathcal{K})$ is Matlab notation
for the matrix $X$ restricted to the columns indexed by $\mathcal{K}$.
Since the coefficients of $H$ are nonnegative, all columns of $X$ live in the conical hull of the
``extreme" columns indexed by $\mathcal{K}$.
The idea of separability was developed by Donoho and Stodden \cite{donoho2003does},
and recent work has produced tractable NMF algorithms by assuming that 
$X$ almost satisfies a separability condition \cite{arora2012computing, bittorf2012factoring, gillis2013robust}.
A matrix $X$ is \emph{noisy $r$-separable} or \emph{near-separable} if
\[
X = X(:, \mathcal{K})H + N,
\]
where $N$ is a noise matrix whose entries are small.
Near-separability means that all data points approximately live in the
conical hull of the extreme columns.

The algorithms for near-separable NMF are typically based on convex geometry and can be described by the same two-step approach:
\begin{enumerate}
\item Determine the extreme columns, indexed by $\mathcal{K}$, and let $W = X(:, \mathcal{K})$.
\item Solve $H= \arg\min_{Y \ge 0} \| X - WY \|$.
\end{enumerate}

The bulk of the literature is focused on the first step.
In Section~\ref{sec:implementation}, we show how to efficiently implement both steps
in a single-pass over the data and provide the details of a MapReduce implementation.

We note that separability is a severe and restrictive assumption.
The tradeoff is that our algorithms are extremely efficient and provably correct under this assumption.
In big data applications, efficiency is at a premium, and this provides some justification for using separability as a tool for exploratory data analysis.
Furthermore, our experiments on real scientific data sets in Section~\ref{sec:applications} under the separability assumption lead to new insights.

\subsection{Alternative NMF algorithms}

There are several approaches to solving Equation~(\ref{eq:nmf}) that do not assume the separability condition.
These algorithms typically employ block coordinate descent, optimizing over $W$ and $H$ while keeping one factor fixed.
Examples include the seminal work by Lee and Seung \cite{seung2001algorithms},
alternating least squares \cite{cichocki2007regularized, kim2013algorithms},
and fast projection-based least squares \cite{kim2008fast}.
Some of these methods are used in MapReduce architectures at scale \cite{liu2010distributed}.

Alternating methods require updating the entire factor $W$ or $H$ after each optimization step.
When one of the factors is large, repeated updated can be prohibitively expensive.
The problem is exacerbated in Hadoop MapReduce \cite{Hadoop2010-021}, where intermediate results are written to disk.
Regardless of the approach or computing platform,
the algorithms are not feasible when the matrices cannot fit in main memory.
In contrast, we show in Sections~\ref{sec:algorithms}~and~\ref{sec:implementation} that the separability
assumption leads to algorithms that do not require updates to large matrices.
This approach is scalable for large tall-and-skinny matrices in big data problems.

\section{Algorithms and dimension \\ reduction for near-separable\\ NMF}
\label{sec:algorithms}

There are several popular algorithms for near-separable NMF,
and they are motivated by convex geometry.
The goal of this section is to show that when $X$ is tall-and-skinny,
we can apply dimension reduction techniques so that established algorithms can execute on $n \times n$ matrices,
rather than the original $m \times n$.
Our new dimension reduction technique in Section~\ref{sec:dimension_reduction} is also motivated by convex geometry.
In Section~\ref{sec:implementation}, we leverage the dimension reduction into scalable algorithms.

\subsection{Geometric algorithms}
\label{sec:geometric}

There are two geometric strategies typically employed for near-separable NMF.
The first deals with conical hulls.
A \emph{cone} $\mathcal{C} \subset \mathbb{R}^m$ is a non-empty convex with
\[
\mathcal{C} = \left\{\sum_{i} \alpha_ix_i | \alpha_i \in \mathbb{R}_+, x_i \in \mathbb{R}^m \right\},
\]
The $x_i$ are generating vectors.
In separable NMF,
\[
X = X(:, \mathcal{K})H
\]
implies that all columns of $X$ lie in the cone generated by the columns indexed by $\mathcal{K}$.
For any $k \in \mathcal{K}$, $\{\alpha X(:, k) | \alpha > 0 \}$ is an \emph{extreme ray} of this cone.
The goal of the XRAY algorithm \cite{kumar2012fast} is to find these extreme rays.
In particular, the greedy variant of XRAY selects the maximum column norm
\[
\arg\max_{j} \| R^TX(:, j) \|_2 / \| X(:, j) \|_2,
\]
where $R$ is a residual matrix that gets updated with each new extreme column.

The second approach deals with convex hulls.
If $D$ is a diagonal matrix with $D_{ii} = \| A(:, i) \|_1$ and $A$ is separable, then
\begin{eqnarray*}
XD^{-1} &=& X(:, \mathcal{K})D(\mathcal{K}, \mathcal{K})^{-1}D(\mathcal{K}, \mathcal{K})HD^{-1} \\
                &=& (XD^{-1})(:, \mathcal{K})\tilde{H}.
\end{eqnarray*}
Thus, $XD^{-1}$ is also separable.
Since the columns are $\ell_1$-normalized,
the columns of $\tilde{H}$ have non-negative entries and sum to one.
In other words, all columns of $XD^{-1}$ are in the convex hull of the columns indexed by $\mathcal{K}$.
The problem is reduced to finding the extreme points of a convex hull.
Popular approaches in the context of NMF include the Successive Projection Algorithm (SPA, \cite{araujo2001successive}) and its generalization \cite{gillis2013fast}.
Another alternative, based on linear programming, is 
Hott Topixx~\cite{bittorf2012factoring}.
As an example of the particulars of one such method, SPA, which we will use in Sections~\ref{sec:synthetic}~and~\ref{sec:applications}, finds extreme points by computing
\[
\arg\max_{j} \| R(:, j) \|_2^2,
\]
where $R$ is a residual matrix related to the data matrix $X$.

In any algorithm, we call the columns indexed by $\mathcal{K}$ \emph{extreme columns}.
The next two subsections are devoted to dimension reduction techniques for finding the extreme columns in the case when $X$ is tall-and-skinny.

\subsection{Orthogonal transformations}
\label{sec:dimension_reduction}

Consider a cone $\mathcal{C} \subset \mathbb{R}^m$ and a nonsingular $m \times m$ matrix $M$.
It is easily shown that $x$ is an extreme ray of $\mathcal{C}$ if and only if $Mx$ is an extreme ray of
$M\mathcal{C} = \{ Mz | z \in \mathcal{C} \}$.
Similarly, for any convex set $\mathcal{S}$, $x$ is an extreme point of $\mathcal{S}$ if and only if $Mx$ is an extreme point of $M\mathcal{S}$.

We take advantage of these facts by applying specific orthogonal transformations as the invertible matrix $M$.
Let $X = Q\tilde{R}$ and $X = U\tilde{\Sigma} V^T$ be the \emph{full} QR factorization and singular value decomposition (SVD) of $X$, so that
$Q$ and $U$ are $m \times m$ orthogonal (and hence nonsingular) matrices.
Then
\[
Q^TX = \begin{pmatrix} R \\ \textbf{0} \end{pmatrix}, \quad U^TX = \begin{pmatrix} \Sigma V^T \\ \textbf{0} \end{pmatrix},
\]
where $R$ and $\Sigma$ are the top $n \times n$ blocks of $\tilde{R}$ and $\tilde{\Sigma}$ and $\textbf{0}$ is an $(m - n) \times n$ matrix of zeroes.
The zero rows provide no information on which columns of $Q^TX$ or $U^TX$ are extreme rays or extreme points.
Thus, we can restrict ourselves to finding the extreme columns of $R$ and $\Sigma V^T$.
These matrices are $n \times n$, and we have significantly reduced the dimension of the problem problem.
In fact, if $X = X(:, \mathcal{K})H$ is a separable representation,
we immediately have separated representations for $R$ and $\Sigma V^T$:
\[
R = R(:, \mathcal{K})H, \quad \Sigma V^T = \Sigma V^T(:, \mathcal{K})H.
\]

We note that, although any invertible transformation preserves extreme columns,
many transformations will destroy the geometric structure of the data.
However, orthogonal transformations are either rotations or reflections,
and they preserve the data's geometry.

This dimension reduction technique is exact when $X$ is $r$-separable,
and the results will be the same for orthogonal transformations $Q^T$ and $U^T$.
This is a consequence of the transformed data having the same separability as the original data.
The SPA and XRAY algorithms briefly described in Section~\ref{sec:geometric} only depend on computing column $2$-norms, which are preserved under orthogonal transformations.
For these algorithms, applying $Q^T$ or $U^T$ preserves the column $2$-norms of the data, and the selected extreme columns are the same.
However, other NMF algorithms do not possess this invariance.
For this reason, we present both of the orthogonal transformations.

Finally, we highlight an important benefit of this dimension reduction technique.
In many applications, the data is noisy and the separation rank ($r$ in Equation~(\ref{eq:nmf}))
is not known \emph{a priori}.
In Section~\ref{sec:compute_H}, we show that the $H$ factor can be computed in the small dimension.
Thus, it is viable to try several different values of the separation rank and pick the best one.
This idea is extremely useful for the applications presented in Section~\ref{sec:applications},
where we do not have a good estimate of the separability of the data.

\subsection{Gaussian projection}
\label{sec:Gaussian}

An alternative dimension reduction technique is random Gaussian projections,
and the idea has been used in hyperspectral unmixing problems \cite{bioucas2011overview}.
In the hyperspectral unmixing literature, the separability is referred to as the \emph{pure-pixel assumption}, and
the random projections are also motivated by convex geometry \cite{boardman1993automating}.
In particular, given a matrix $G \in \mathbb{R}^{m \times k}$ with Gaussian i.i.d.~entries,
the extreme columns of $X$ are taken as the extreme columns of $G^TX$, which is of dimension $k \times n$.
The algorithm assumes that the columns of $X$ are normalized to sum to one.
In other words,
\[
X(:, i) \leftarrow X(:, i) / ||X(:, i)||_1, \quad i = 1, \ldots, n
\]
In Section~\ref{sec:colnorms}, we show how to run the algorithm in one pass over the data matrix, even if the columns are not normalized.
Recent work shows that when $X$ is $r$-separable and $k = O(r \log r)$, then all of the extreme columns are found with high probability \cite{damle2014randomized}.
The extreme column selection is simple:
for each row of $G^TX$, the indices of the minimum and maximum entries are added to the set $\mathcal{K}$.
The Gaussian transformation translates cleanly to streaming and MapReduce algorithms.
We provide an implementation in Section~\ref{sec:GP_impl}.

\subsection{Computing H}
\label{sec:compute_H}

Selecting the extreme columns indexed by $\mathcal{K}$ completes one half of the NMF factorization in Equation~(\ref{eq:nmf}).
How do we compute $H$?
We want
\[
H = \arg\min_{Y \in \mathbb{R}^{r \times n}_+} \| X - X(:, \mathcal{K})Y \|^2
\]
for some norm.
Choosing the Frobenius norm results in a set of $n$ nonnegative least squares (NNLS) problems:
\[
H(:, i) = \arg\min_{y \in \mathbb{R}^r_+} \| X(:, \mathcal{K})y - X(:, i) \|_2^2, \quad i = 1, \ldots, n.
\]
Let $X = Q\tilde{R}$ with $R$ the upper $n \times n$ block of $\tilde{R}$.
Then
\begin{eqnarray*}
H(:, i) &=&   \arg\min_{y \in \mathbb{R}^r_+} \| X(:, \mathcal{K})y - X(:, i) \|_2^2 \\
           &=& \arg\min_{y \in \mathbb{R}^r_+} \| Q^T\left(X(:, \mathcal{K})y - X(:, i)\right) \|_2^2 \\
           &=& \arg\min_{y \in \mathbb{R}^r_+} \| R(:, \mathcal{K})y - R(:, i) \|_2^2
\end{eqnarray*}

Thus, we can solve the NNLS problem with matrices of size $n \times n$.
After computing just the small $R$ factor from the QR factorization,
we can compute the entire nonnegative matrix factorization by working with matrices of size $n \times n$.
Analogous results hold for the SVD, where we replace $Q$ by $U$, the left singular vectors.
In Section~\ref{sec:implementation}, we show that these computations are simple and scalable.
Since $m \gg n$, computations on $O(n^2)$ data are fast, even in serial.
Finally, note that we can also compute the residual in this reduced space, i.e.:
\[
\min_{y \in \mathbb{R}^n_+} \| X(:, \mathcal{K})y - X(:, i) \|_2^2 = \min_{y \in \mathbb{R}^n_+} \| R(:, \mathcal{K})y - R(:, i) \|_2^2.
\]
This simple fact is significant in practice.
When there are several candidate sets of extreme columns $\mathcal{K}$,
the residual error for each set can be computed quickly.
In Section~\ref{sec:applications}, we compute many residual errors for different sets $\mathcal{K}$
in order to choose an optimal separation rank.

We have now shown how to use dimension reduction techniques for tall-and-skinny matrix data in near-separable NMF algorithms.
Following the same strategy as many NMF algorithms, we first compute extreme columns and then solve for the coefficient matrix $H$.
Fortunately, once the upfront cost of the orthogonal transformation is complete, both steps can be computed using $O(n^2)$ data.

\section{Implementation}
\label{sec:implementation}

Remarkably, when the matrix is tall-and-skinny,
we need only one pass over the data matrix and a MapReduce implementation suffices to achieve optimal communication. 
While all of the algorithms use sophisticated computation, these routines are only ever invoked with matrices of size $n \times n$.
Thus, we get extremely scalable implementations.

\subsection{TSQR and R-SVD}
\label{sec:tsqr}

The thin QR factorization of an $m \times n$ real-valued matrix $A$ with $m > n$ is
\[
A = QR
\]
where $Q$ is an $m \times n$ orthogonal matrix and $R$ is an $n \times n$ upper triangular matrix.
This is precisely the factorization we need in Section~\ref{sec:algorithms}.
For our purposes, $Q^T$ is applied implicitly, and we only need to compute $R$.
When $m \gg n$, communication-optimal algorithms for computing the factorization are referred to as TSQR \cite{demmel2008communication}.
TSQR is implemented in several environments, including
MapReduce \cite{benson2013direct, constantine2011tall},
distributed memory MPI \cite{demmel2008communication},
GPUs \cite{anderson2011communication},
and grid computing \cite{agullo2010qr}.
All of these methods avoid computing $A^TA$ and hence are numerically stable.
The dimension reduction techniques from Section~\ref{sec:algorithms} are independent of the platform.
However, as explained in Section~\ref{sec:mapreduce}, we use MapReduce to target data computations.

The thin SVD factorization used in Section~\ref{sec:dimension_reduction} is a small extension of the thin QR factorization.
The thin SVD is
\[
A = U\Sigma V^T
\]
where $U$ is $m \times n$ and orthogonal, $\Sigma$ is diagonal with decreasing, nonnegative diagonal entries, and $V$ is $n \times n$ and orthogonal.
Let $A = QR$ be the thin QR factorization of $A$ and $R = U_R\Sigma V^T$ be the SVD of $R$.
Then
\[
A = (QU_R)\Sigma V^T = U\Sigma V^T.
\]
The matrix $U = QU_R$ is $m \times n$ and orthogonal, so this is the thin SVD factorization of $A$.
The dimension of $R$ is $n \times n$, so computing its SVD takes $O(n^3)$ floating point operations (flops), a trivial cost when $n$ is small.
When $m \gg n$, this method for computing the SVD is called the $R$-SVD \cite{chan1982improved}.
Both TSQR and $R$-SVD require $O(mn^2)$ flops.
However, the dominant cost is communication,
which we analyze in Section~\ref{sec:communication}.

\subsection{Gaussian projection}
\label{sec:GP_impl}

\begin{algorithm}[t]
\caption{MapReduce Gaussian Projection for NMF}
   \label{alg:GP}
   \begin{algorithmic}
      \Function{Map}{key $k$, matrix row $x_i^T$}
      \State Sample column vector $g_i \sim N(0, I_k)$.
      \For{each row $r_k$ of $g_ix_i^T$}
          \State emit $(k, r_k)$
      \EndFor
      \EndFunction
      \Function{Reduce}{key $k$, matrix rows $\langle r_k \rangle$}
       \State emit $(k, \text{sum}(\langle r_k \rangle))$
      \EndFunction      
   \end{algorithmic}
\end{algorithm}

For implementing the Gaussian projection in Section~\ref{sec:Gaussian},
we assume that we can quickly sample i.i.d.~Gaussians in parallel.
Under this assumption, the transformation is easily implemented in MapReduce.
For each row of the data matrix $x_i^T$, the map function computes the outer product $g_ix_i^T$,
where $g_i$ consists of $n$ sampled Gaussians.
The map function emits a key-value pair for each row of this outer product, where the key is the row number.
The reduce function simply sums all rows with the same key.
Algorithm~\ref{alg:GP} contains the functions.
In theory and practice, all of the outer products on a single map process are summed before emitting key-value pairs.
The function performing the aggregation is called a combiner.

In practice, we can only generate i.i.d.~Gaussians on a single processor.
While there may be small correlations in samples across processors,
this does not affect the performance of the algorithm in practice (see Section~\ref{sec:synthetic}).

\subsection{Column normalization}\label{sec:colnorms}

The convex hull algorithms from Section~\ref{sec:geometric} and the Gaussian projection algorithm from Section~\ref{sec:Gaussian} require the columns of the data matrix $X$ to be normalized.
A naive implementation of the column normalization for the convex hull algorithms in a MapReduce or streaming environment would do the following:
\begin{enumerate}
\item Read $X$ and compute the column norms.
\item Read $X$, normalize the columns, and write the normalized data to disk.
\item Use TSQR on the normalized matrix.
\end{enumerate}

This requires reading the data matrix twice times and writing $O(mn)$ data to disk once to just normalize the columns.
The better approach is a single step:
\begin{enumerate}
\item Use TSQR on the unnormalized data $X$ and simultaneous compute the column norms of $X$.
\end{enumerate}

Let $D$ be the diagonal matrix of column norms.  Note that
\[
X = QR \to XD^{-1} = Q(RD^{-1}).
\]
The matrix $\hat{R} = RD^{-1}$ is upper triangular, so $Q\hat{R}$ is the thin QR factorization of the column-normalized data.
This approach reads the data once and only writes $O(n^2)$ data.
The same idea applies to Gaussian projection:
\[
G^T(XD^{-1}) = (G^TX)D^{-1}.
\]
Thus, our algorithms only need to read the data matrix once in all cases.

\subsection{MapReduce motivation}
\label{sec:mapreduce}

For the experiments in this paper, we use a MapReduce implementations for the NMF algorithms presented in Section~\ref{sec:geometric} using the dimension reduction techniques in Section~\ref{sec:dimension_reduction}.
Our central computational kernel is the tall-and-skinny QR factorization (TSQR), which has been optimized on several architectures (see the references in the Section~\ref{sec:tsqr}).
Thus, our ideas in this paper are not restricted to MapReduce architectures.
That being said, MapReduce remains a popular framework in data-intensive computing for several reasons.
First, many large datasets are already warehoused in MapReduce clusters.
Having algorithms that run on the cluster eliminate the need to transfer data to another computer.
The algorithms in this paper need only one pass over the data.
Since running time is dominated by the cost of loading data from disk to main memory,
the time to transfer data can take as long as simply running the algorithm directly on the cluster.
Second, systems like Hadoop \cite{Hadoop2010-021} and Spark \cite{zaharia2012resilient} typically manage the distributed file input-output routines and communication collectives.
This significantly reduces the software development cycle.
Finally, many MapReduce implementations provide transparent fault tolerance.

\subsection{Communication costs for NMF on MapReduce}
\label{sec:communication}

There are two communication costs that we analyze for MapReduce.
The first is the time to read the input data.
In Hadoop, data is stored on disk and loading the data is frequently the dominant cost in numerical algorithms.
The second is the time spent shuffling data.
This is roughly measured by the number and size of the key-value pairs sorted in the shuffle step.
Current implementations of TSQR and $R$-SVD in MapReduce can compute $R$ or $\Sigma V^T$ in a a single MapReduce iteration \cite{benson2013direct}.
For the dimension reduction, the data matrix only needs to be read once.
Although algorithms such as Hott Topixx, SPA, and Gaussian projection require normalized columns,
we showed that the column norms can be computed at the same time as TSQR (see Section~\ref{sec:colnorms}).
For Gaussian projection, 
we cannot compute the factor $H$ in the same projected space.
To remedy this, we combine TSQR with the Gaussian projection in a single pass over the data.
Following this initial computation, the $H$ matrix is computed as in Section~\ref{sec:compute_H}.

The map processes in the MapReduce implementations for TSQR, $R$-SVD, and Algorithm~\ref{alg:GP}
emit $O(n \cdot \#(\text{map tasks}))$ keys to the shuffle stage (one for each row of the reduced matrix). 
The key-value pairs are $O(n)$ in length---each pair represents a partial row sum of the resultant $n \times n$ matrix.
For tall-and-skinny matrices, $n$ may as well be considered a constant as it is often incredibly small. Thus, our communication is optimal.

\section{Testing on synthetic matrices}
\label{sec:synthetic}

\begin{figure}[tb]
\centering
\includegraphics[height=3in]{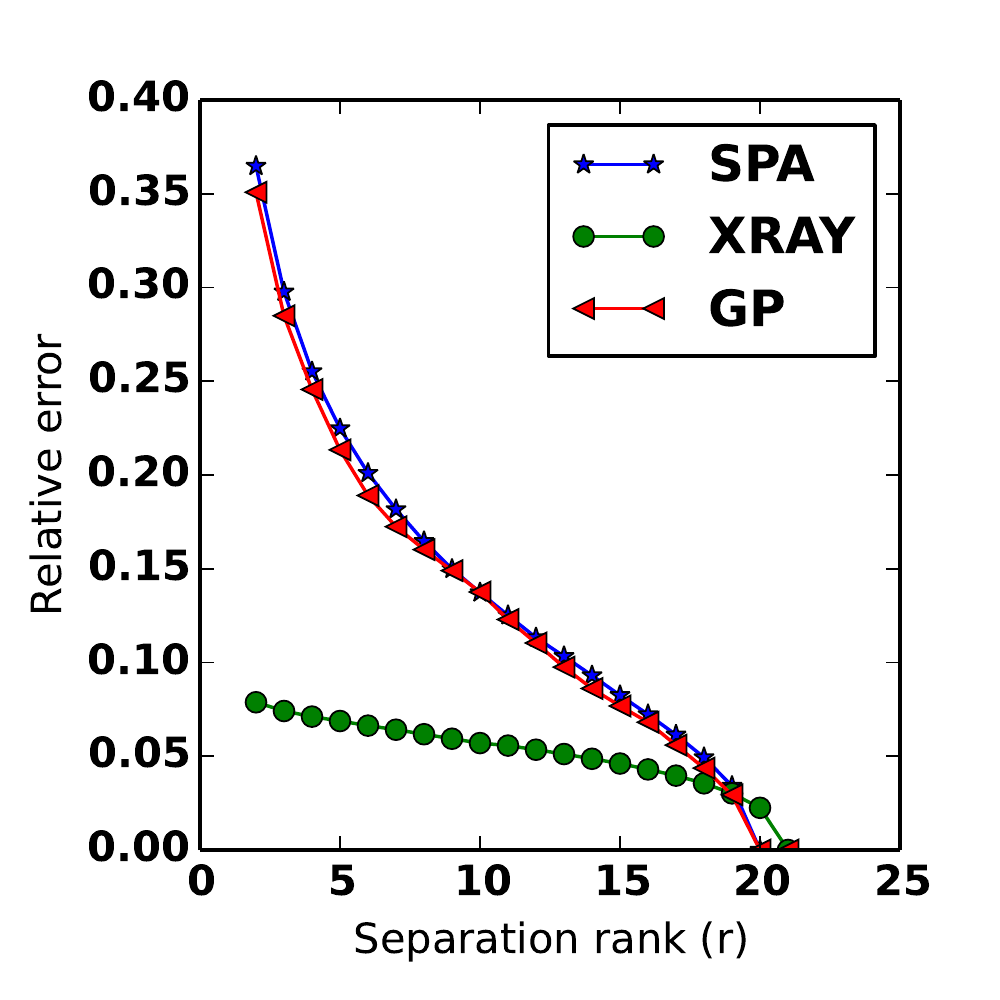}
\captionskip 
\caption{
Relative error in the separable factorization as a function of nonnegative rank ($r$) for the three algorithms.
The matrix was synthetically generated to be separable.
SPA and GP capture all of the true extreme columns when $r = 20$ (where the residual is zero).
Since we are using the greedy variant of XRAY, it takes $r = 21$ to capture all of the extreme columns.
}
\label{fig:synth_exact_residuals}
\end{figure}

\begin{figure}[tb]
\centering
\includegraphics[height=3in]{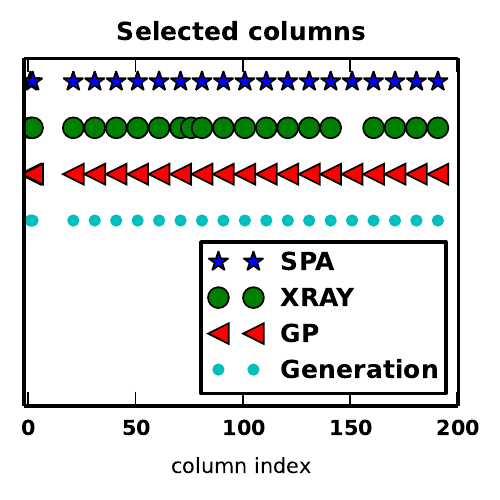}
\captionskip 
\caption{
First $20$ extreme columns selected by SPA, XRAY, and GP along with the true columns used in the synthetic matrix generation.
A marker is present for a given column index if and only if that column is a selected extreme column.
SPA and GP capture all of the true extreme columns.
Since we are using the greedy variant of XRAY, it does select all of the true extreme columns (the columns marked Generation).
}
\label{fig:synth_exact_cols}
\end{figure}

\begin{figure*}[tb]
\centering
\includegraphics[scale=0.7]{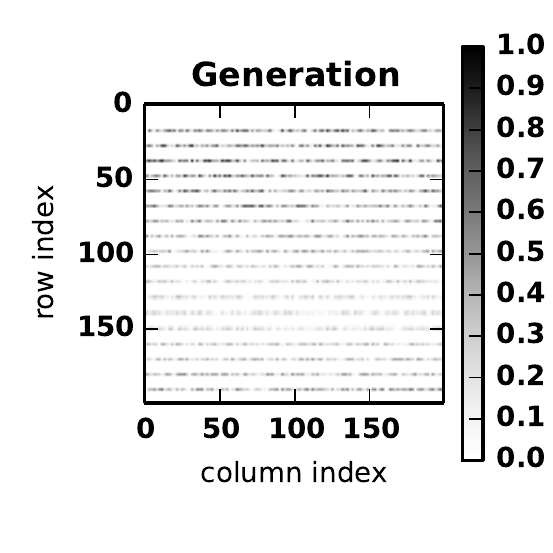}
\includegraphics[scale=0.7]{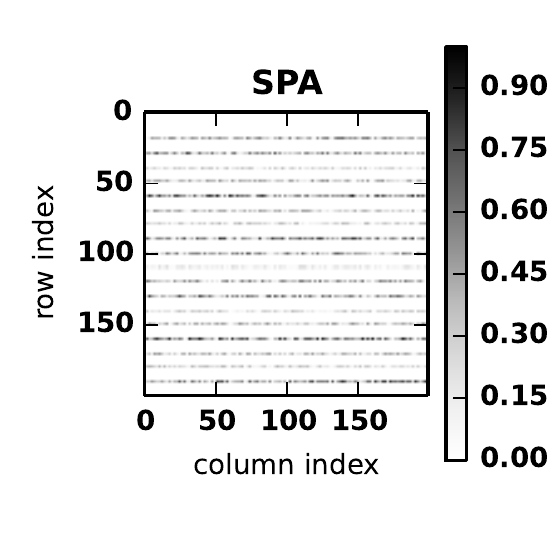}
\includegraphics[scale=0.7]{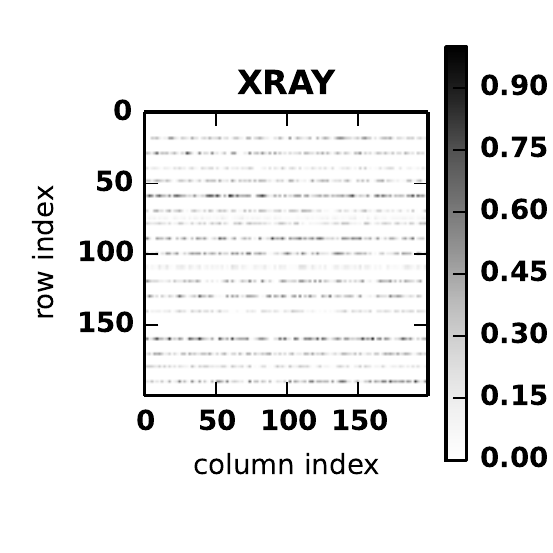}
\includegraphics[scale=0.7]{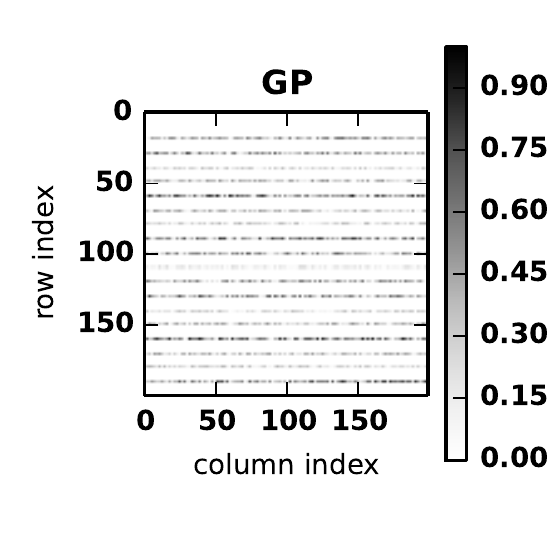}
\captionskip 
\caption{
Coefficient matrix $H$ found by the algorithms when $r = 20$, where the test matrix was synthetically generated to have separation rank $20$.
The synthetically generated matrix is in the left plot.
Each algorithm captures similar coefficients.
The entries in $H$ for the original matrix were generated from uniform random variables, so there is no visible structure in the coefficients.
In real applications, the coefficients have significant structure (see Figures~\ref{fig:heat_coeffs}~and~\ref{fig:cells_coeffs}).
}
\label{fig:synth_exact_coeffs}
\end{figure*}

In this section, we test our dimension reduction techniques on tall-and-skinny matrices that are synthetically generated to be separable or near-separable.
All experiments were conducted on a 10-node, 40-core MapReduce cluster at Stanford's Institute for Computational and Mathematical Engineering (ICME). Each node has 6 2-TB disks, 24 GB of RAM, and a single Intel Core i7-960 3.2 GHz processor. They are connected via Gigabit ethernet.
We test the following three algorithms:
\begin{enumerate}
\item Dimension reduction with the SVD followed by SPA.
\item Dimension reduction with the SVD followed by the greedy variant of the XRAY algorithm.
The greedy method is not exact in the separable case but works well in practice \cite{kumar2012fast}.
\item Gaussian projection (GP) as described in Section~\ref{sec:Gaussian}.
\end{enumerate}

Using our dimension reduction technique, all three algorithms require only one pass over the data.
The algorithms were selected to be a representative set of the approaches in the literature,
and we will refer to the three algorithms as SPA, XRAY, and GP.
As discussed in Section~\ref{sec:dimension_reduction}, the choice of QR or SVD does not matter for these algorithms (although it may matter for other NMF algorithms).
Thus, we only consider the SVD transformation in the subsequent numerical experiments.

We generate a separable matrix $X$ with $m = 200$ million rows and $n = 200$ columns.
The nonnegative rank ($r$ in Equation~(\ref{eq:nmf})) is $20$.
We generated the separable matrix by
\[
X := W\begin{pmatrix} I_{r} & H^{\prime} \end{pmatrix} \Pi,
\]
where $H^{\prime}$ is a $r \times (n - r)$ and $W$ is a $m \times r$ matrix with entries generated from a Uniform $[0, 1]$ distribution.
The permutation matrix $\Pi$ swaps columns $i$ and $10i$, $i = 2, \ldots, r-1 = 19$.
In other words, the extreme columns of $X$ are indexed by $0, 1, 10, 20, \ldots, 190$.
The matrix occupies 286 GB of storage in the Hadoop Distributed File System (HDFS).

Figure~\ref{fig:synth_exact_residuals} shows the relative residual error as a function of the separation rank $r$.
The relative error is
\[
\| X - X(:, \mathcal{K})H \|_F^2 / \| X \|_F^2.
\]
As $r$ approaches $20$, the true separation rank, the relative error converges to 0.
Figure~\ref{fig:synth_exact_cols} shows the columns selected when $r = 20$.
All algorithms except XRAY select the correct columns.
Since we use the greedy variant of XRAY, it is not guaranteed to select the correct columns.
However, all $20$ extreme columns are in the first $21$ extreme columns selected by XRAY.
Figure~\ref{fig:synth_exact_coeffs} shows the coefficient matrix $H$ computed with each algorithm
when $r = 20$.
We see that the NNLS solver successfully recovers the correct coefficients.

\begin{figure}[tb]
\centering
\includegraphics[height=3in]{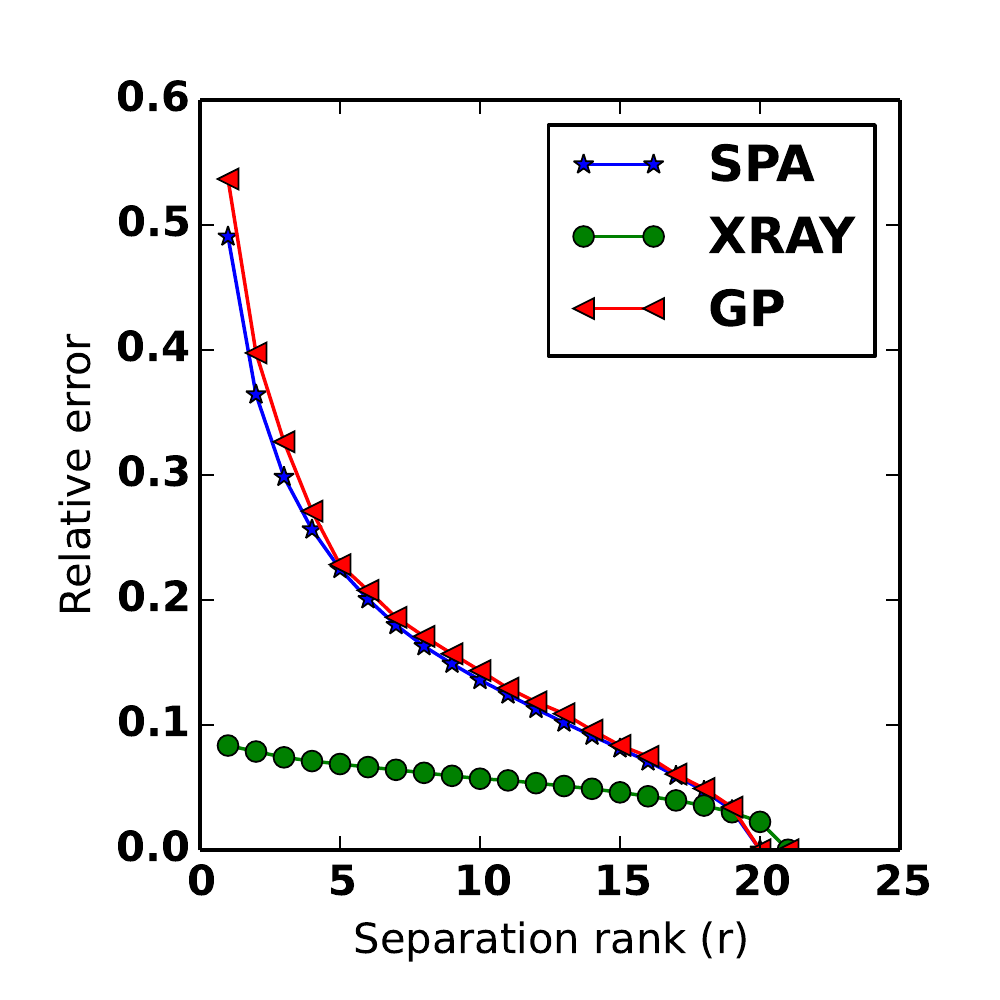}
\captionskip 
\caption{
Relative error in the separable factorization as a function of nonnegative rank ($r$) for the three algorithms.
The matrix was synthetically generated to be nearly $r$-separable (exactly separable with small, additive noise).
The results agree with the noiseless case (see Figure~\ref{fig:synth_exact_residuals}).
}
\label{fig:synth_noisy_residuals}
\end{figure}

\begin{figure}[tb]
\centering
\includegraphics[height=3in]{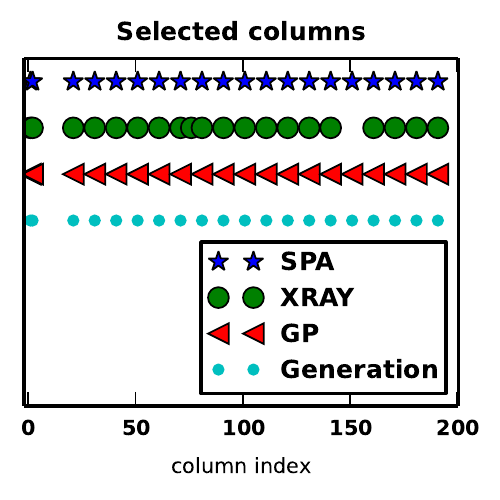}
\captionskip 
\caption{
First $20$ extreme columns selected by SPA, XRAY, and GP along with the true columns used in the synthetic matrix generation.
In this case, the data is noisy and the matrix is nearly $r$-separable.
However, the results are the same as in the noiseless case (see Figure~\ref{fig:synth_exact_cols}).
}
\label{fig:synth_noisy_cols}
\end{figure}

\begin{figure*}[tb]
\centering
\includegraphics[scale=1]{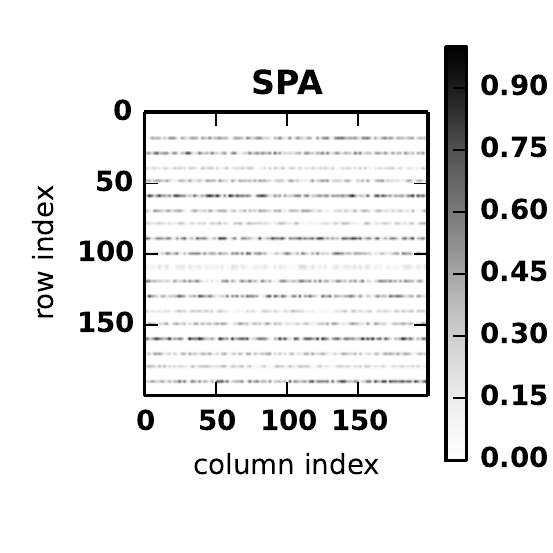}
\includegraphics[scale=1]{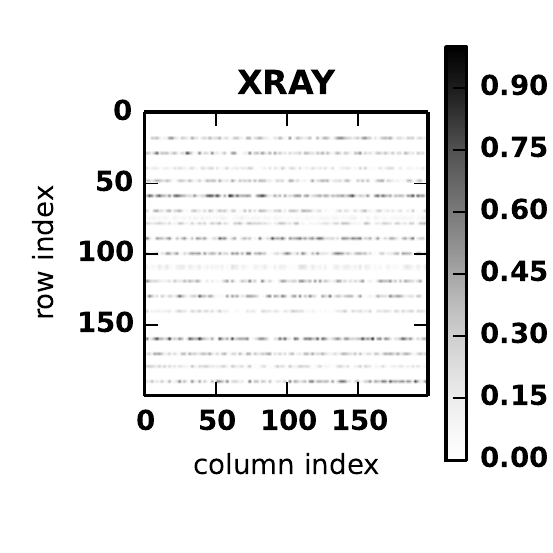}
\includegraphics[scale=1]{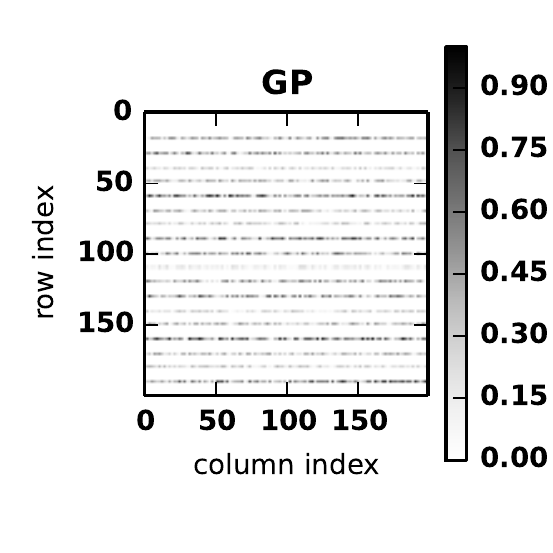}
\captionskip 
\caption{
Coefficient matrix $H$ for SPA, XRAY, and GP for the synthetic near-separable matrix when $r = 20$.
Overall, the coefficients are similar to those in Figure~\ref{fig:synth_exact_coeffs}, where there is no noise.
Again, the coefficients are generated from uniform random variables, so there is no structure in the coefficients themselves.
}
\label{fig:synth_noisy_coeffs}
\end{figure*}

We also test our algorithm with a near-separable matrix:
\[
X := W\begin{pmatrix} I_{r} & H^{\prime} \end{pmatrix} \Pi + N,
\]
The matrices $W$, $H$, and $\Pi$ are the same as the above experiment.  The entries of $N$ are chosen uniformly on $[0$, 1e-3$]$.
Figures~\ref{fig:synth_noisy_residuals}~\ref{fig:synth_noisy_cols},~and~\ref{fig:synth_noisy_coeffs} show the relative errors for varying separation ranks, the columns selected with $r = 20$, and the coefficient matrices $H$ for each algorithm.
The results are identical to those in the separable case.

\section{Applications}
\label{sec:applications}

We now test our algorithms and implementations on scientific data sets,
using the same algorithms and compute system configuration described in Section~\ref{sec:synthetic}.
The data are nonnegative, but we do not know \emph{a priori} that the data is separable.

\subsection{Heat transfer simulation data}
\label{sec:heat}

The heat transfer simulation data contains the simulated heat in a high-conductivity stainless steel block with a low-conductivity foam bubble inserted in the block \cite{constantine2013model}.
Each column of the matrix corresponds to simulation results for a foam bubble of a different radius.
Several simulations for random foam bubble locations are included in a column.
Each row corresponds to a three-dimensional spatial coordinate, a time step, and a bubble location.
An entry of the matrix is the temperature of the block at a single spatial location, time step, bubble location, and bubble radius. The matrix is constructed such that columns near $64$ have far more variability in the data -- this is then responsible for additional ``rank-like'' structure. Thus, we would intuivitely expect the NMF algorithms to select additional columns closer to the end of the matrix. (And indeed, this is what we will see shortly.)
In total, the matrix has approximately 4.9 billion rows and 64 columns and occupies a little more than 2 TB on HDFS.
The data set is publicly available.\footnote{\url{https://www.opensciencedatacloud.org/publicdata/}}

\begin{figure}[tb]
\centering
\includegraphics[height=3in]{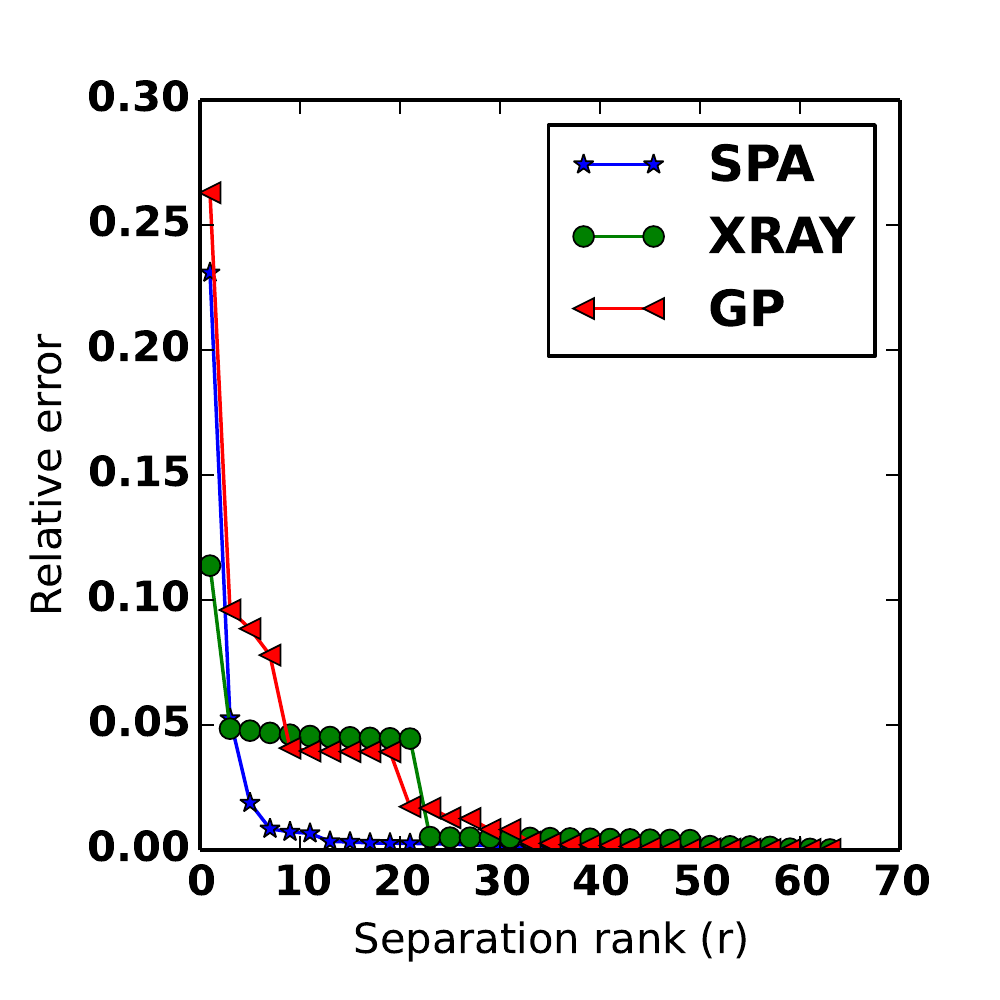}
\captionskip 
\caption{
Relative error in the separable factorization as a function of nonnegative rank ($r$) for the heat transfer simulation data.
Because of our dimension reduction technique, we can quickly compute the residuals and choose a value of $r$ that makes sense for the data.
In this case, we choose $r = 10$, as there is an ``elbow" in the GP curve there.
}
\label{fig:heat_residuals}
\end{figure}

Figure~\ref{fig:heat_residuals} shows the residuals for varying separation ranks.
Even a small separation rank ($r = 4$) results in a small residual.
SPA has the smallest residuals, and XRAY and GP are comparable.
An advantage of our projection method is that we can quickly test many values of $r$.
For the heat transfer simulation data, we choose $r = 10$ for further experiments.
This value of $r$ is near an ``elbow" in the residual plot for the GP curve.

\begin{figure}[tb]
\centering
\includegraphics[height=3in]{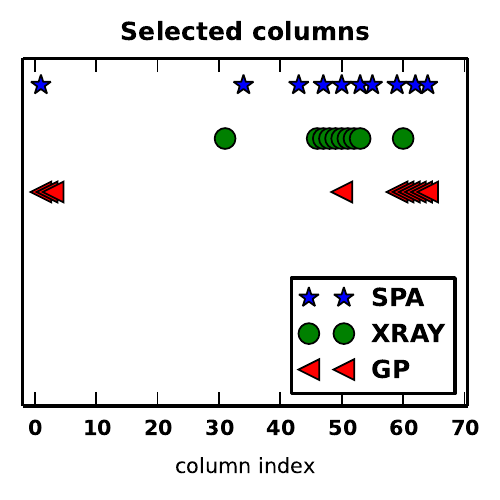}
\captionskip 
\caption{
First $10$ extreme columns selected by SPA, XRAY, and GP for the heat transfer simulation data.
The separation rank $r = 10$ was chosen based on the residual curves in Figure~\ref{fig:heat_residuals}.
For the heat transfer simulation data, the columns with larger indices are more extreme.
However, the algorithms still select different extreme columns.
}
\label{fig:heat_cols}
\end{figure}

\begin{figure*}[tb]
\centering
\includegraphics[scale=1]{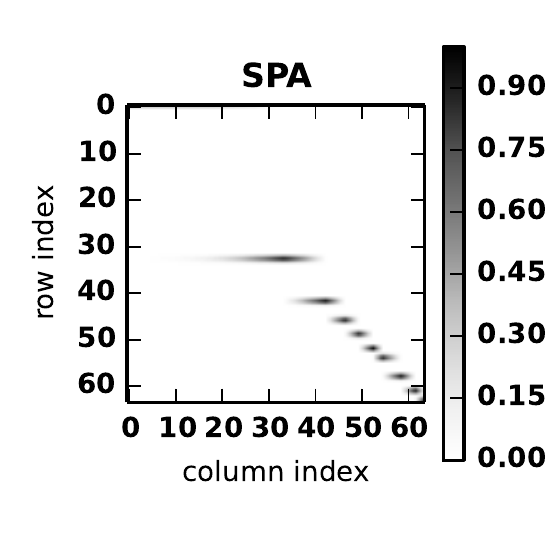}
\includegraphics[scale=1]{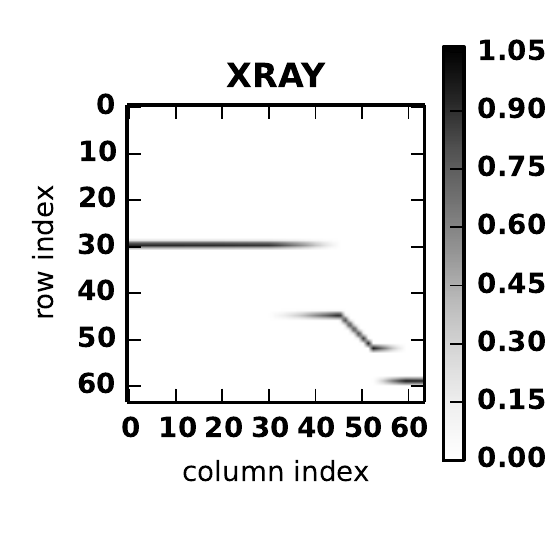}
\includegraphics[scale=1]{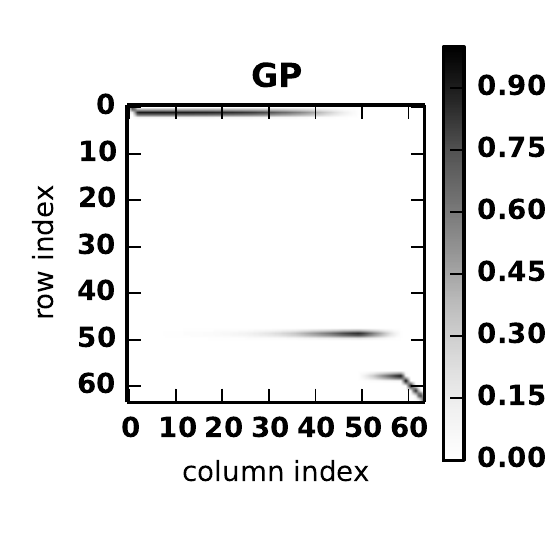}
\captionskip 
\caption{
Coefficient matrix $H$ for SPA, XRAY, and GP for the heat transfer simulation data when $r = 10$.
In all cases, the non-extreme columns are conic combinations of two of the selected columns, i.e., each column in $H$ has at most two non-zero values.
Specifically, the non-extreme columns are conic combinations of the two extreme columns that ``sandwich" them in the matrix.
See Figure~\ref{fig:heat_combinations} for a closer look at the coefficients.
}
\label{fig:heat_coeffs}
\end{figure*}

\begin{figure}[tb]
\centering
\includegraphics[height=3in]{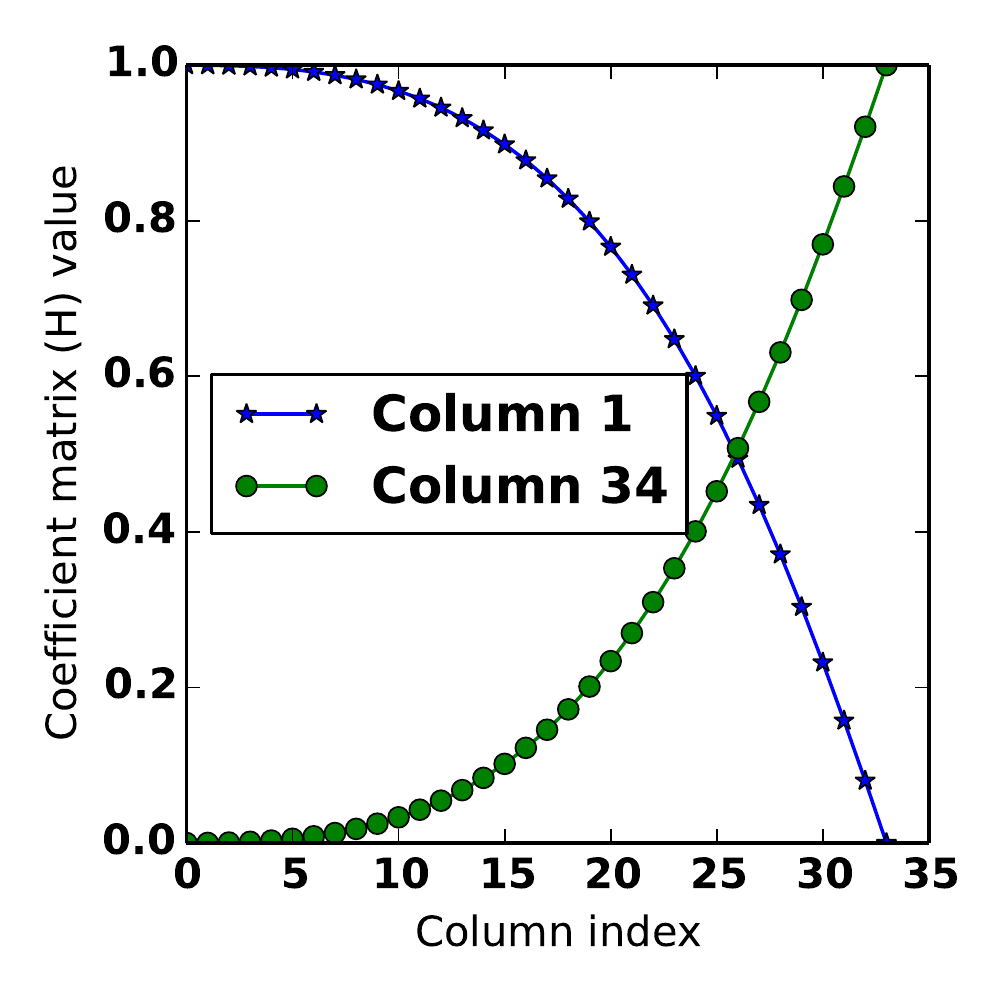}
\captionskip 
\caption{
Value of $H$ matrix for columns $1$ through $34$ for the SPA algorithm on the heat transfer simulation data matrix with separation rank $r = 10$.
Columns $1$ and $34$ were selected as extreme columns by the algorithm,
while columns $2$ through $33$ were not.
The two curves show the value of the matrix $H$ in rows $1$ and $34$ for many columns.
For these columns of $H$, the value is negligible for other rows.
}
\label{fig:heat_combinations}
\end{figure}

Figure~\ref{fig:heat_cols} shows the columns selected by each algorithm.
Columns five through $30$ are not extreme in any algorithm.
Both SPA and GP select at least one column in indices one through four.
Columns $41$ through $64$ have the highest density of extreme columns for all algorithms.
Although the extreme columns are different for the algorithms, the coefficient matrix $H$ exhibits remarkably similar characteristics in all cases.
Figure~\ref{fig:heat_coeffs} visualizes the matrix $H$ for each algorithm.
Each non-extreme column is expressed as a conic combination of only two extreme columns.
In general, the two extreme columns corresponding to column $i$ are
\[
j_1 = \arg\max \{ j \in \mathcal{K} | j < i \}, \quad j_2 = \arg\min \{ j \in \mathcal{K} | j > i \}.
\]
In other words, a non-extreme column is a conic combination of the two extreme columns that ``sandwich" it in the data matrix.
Furthermore, when the index $i$ is closer to $j_1$, the coefficient for $j_1$ is larger and the coefficient for $j_2$ is smaller.
This phenomenon is illustrated in Figure~\ref{fig:heat_combinations}.

\subsection{Flow cytometry}
\label{sec:flow}

The flow cytometry (FC) data represents abundances of fluorescent
molecules labeling antibodies that bind to specific targets on the
surface of blood cells. The phenotype and function of individual cells
can be identified by decoding these label combinations. The analyzed
data set contains measurements of 40,000 single cells. The measurement
fluorescence intensity conveying the abundance information were collected
at five different bands corresponding to the FITC, PE, ECD, PC5, and PC7
fluorescent labels tagging antibodies against CD4, CD8, CD19, CD45, and
CD3 epitopes.

The results are represented as the data matrix $A$ of size $40,000
\times 5$. Our interest in the presented analysis was to study pairwise
interactions in the data (cell vs. cell, and marker vs. marker). Thus,
we are interested in the matrix $X = A \otimes A$, the Kronecker product
of $A$ with itself. Each row of $X$ corresponds to a pair of cells and
each column to a pair of marker abundance values. $X$ has dimension
$40,000^2 \times 5^2$ and occupies 345 GB on HDFS.

\begin{figure}[tb]
\centering
\includegraphics[height=3in]{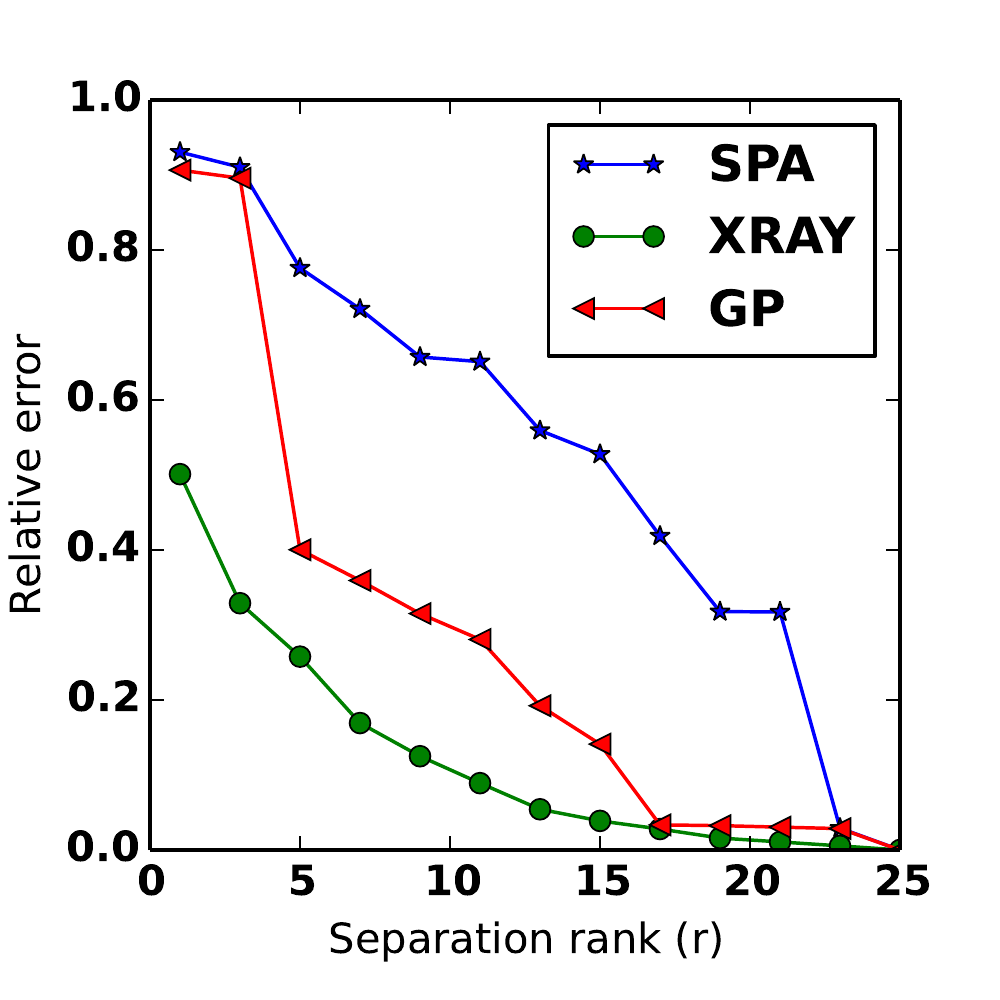}
\captionskip 
\caption{
Relative error in the separable factorization as a function of nonnegative rank ($r$) for the flow cytometry data.
Our dimension reduction technique allows us to compute the errors quickly and choose a value of $r$ that makes sense for the data.
In this case, we choose $r = 16$ since the XRAY GP curve levels for larger $r$.
}
\label{fig:cells_residuals}
\end{figure}

\begin{figure}[tb]
\centering
\includegraphics[height=3in]{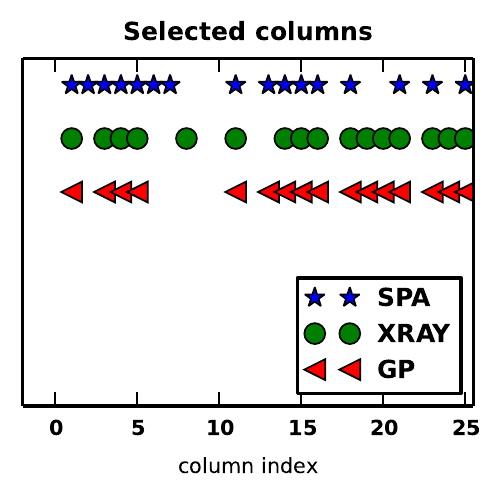}
\captionskip 
\caption{First $16$ extreme columns selected by SPA, XRAY, and GP for the flow cytometry data.
The separation rank of $16$ was selected based on the residual curves in Figure~\ref{fig:cells_residuals}.
}
\label{fig:cells_cols}
\end{figure}

Figure~\ref{fig:cells_residuals} shows the residuals for the three
algorithms applied to the FC data for varying values of the separation
rank. In contrast to the heat transfer simulation data, the relative
errors are quite large for small $r$. In fact, SPA has large relative
error until nearly all columns are selected ($r = 22$).
Figure~\ref{fig:cells_cols} shows the columns selected when $r = 16$.
XRAY and GP only disagree on one column. SPA chooses different columns,
which is not surprising given the relative residual error.
Interestingly, the columns involving the second marker defining the
phenotype (columns 2, 6, 7, 8, 9, 10, 12, 17, 22) are underrepresented
in all the choices. This suggests that the information provided by the
second marker may be redundant. In biological terms, it may indicate
that the phenotypes of the individual cells can be inferred from a
smaller number of markers. Consequently, this opens a possibility that
in modified experimental conditions, the FC researchers may omit this
particular label, and still be able to recover the complete phenotypic
information. Owing to the preliminary nature of these studies, a more
in-depth analysis involving multiple similar blood samples would be
desirable in order to confirm this hypothesis.

\begin{figure*}[tb]
\centering
\includegraphics[scale=1]{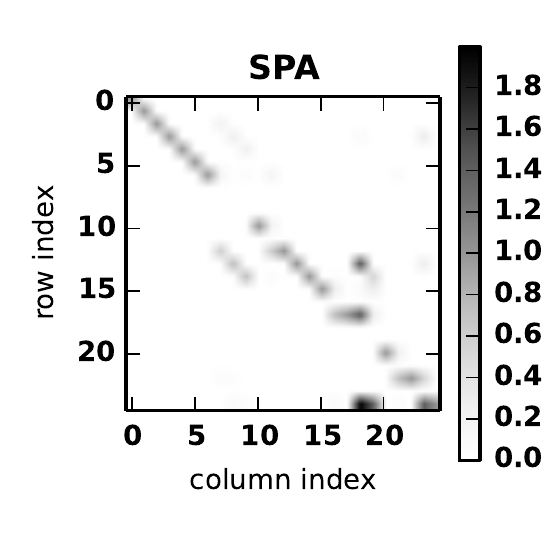}
\includegraphics[scale=1]{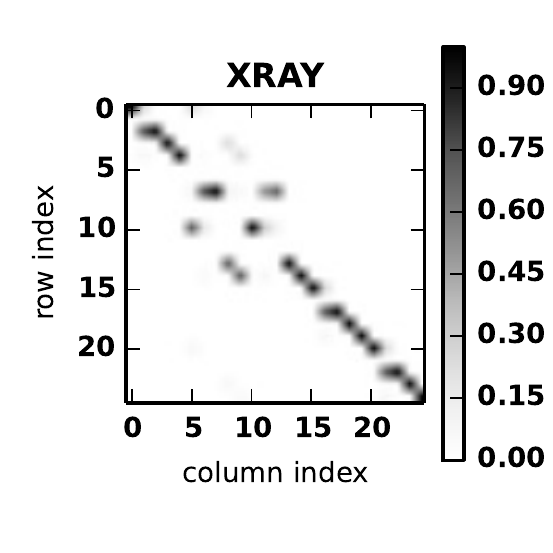}
\includegraphics[scale=1]{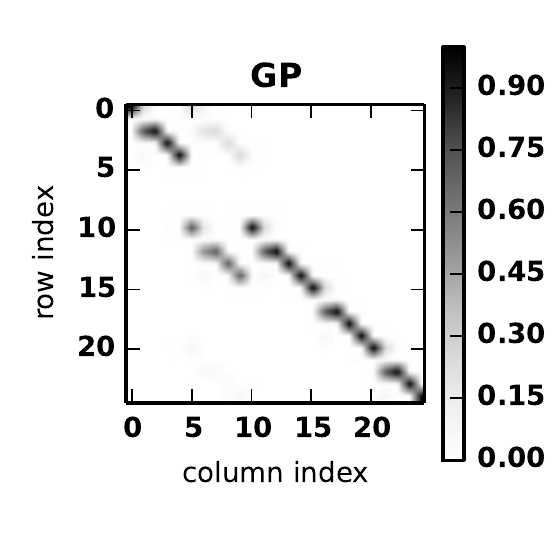}
\captionskip 
\caption{
Coefficient matrix $H$ for SPA, XRAY, and GP for the flow cytometry data when $r = 16$.
The coefficients tend to be clustered near the diagonal.
This is remarkably different to the coefficients for the heat transfer simulation data in Figure~\ref{fig:heat_cols}.
}
\label{fig:cells_coeffs}
\end{figure*}

Finally, Figure~\ref{fig:cells_coeffs} shows the coefficient matrix $H$.
The coefficients are larger on the diagonal, which means that
the non-extreme columns are composed of nearby extreme columns in the matrix.

\section{Discussion}

We have shown how to efficiently compute nonnegative matrix factorizations for near-separable tall-and-skinny matrices.
Our main tool was TSQR, and our algorithms only needed to read the data matrix once.
By reducing the dimension of the problem, we can easily compute the efficacy of factorizations for several values of the separation rank $r$.
With these tools, we have computed the largest separable nonnegative matrix factorizations to date.
Furthermore, our algorithms provide new insights into massive scientific data sets.
The coefficient matrix $H$ exposed structure in the results of heat transfer simulations.
Extreme column selection in flow cytometry showed that one of the labels used in measurements may be redundant.
In future work, we would like to analyze additional large-scale scientific data sets.
We also plan to test additional NMF algorithms.

The practical limits of our algorithm are imposed by the tall-and-skinny requirement where we assume that it is \emph{easy} to manipulate $n \times n$ matrices. The examples we explored used up to $200$ columns and we have explored regimes up to $5000$ columns in prior work~\cite{constantine2011tall}. A rough rule of thumb is that our implementations should be possible as long as an $n \times n$ matrix fits into main memory. This means that implementations based on our work will scale up to $30,000$ columns on machines with more than $8$ GB of memory; although at this point communication begins to dominate. Solving these problems with more columns is a challenging opportunity for the future.

\section{Acknowledgements}

Austin R. Benson is supported by an Office of Technology Licensing Stanford Graduate Fellowship.
Jason D. Lee is supported by an Office of Technology Licensing Stanford Graduate Fellowship and a National Science Foundation Graduate Research Fellowship.
We thank Anil Damle and Yuekai Sun for helpful discussions.

\fontsize{8}{9}\selectfont
\bibliographystyle{abbrv}
\bibliography{bibliography}  
%

\end{document}